\documentclass[11pt]{article}

\usepackage[utf8]{inputenc}
\usepackage[T1]{fontenc}
\usepackage{lmodern}  
\usepackage[margin=1in]{geometry}
\usepackage{microtype}
\usepackage{graphicx}
\usepackage{booktabs}
\usepackage{amsfonts}
\usepackage{xcolor}
\usepackage[numbers]{natbib}
\usepackage{hyperref}
\hypersetup{colorlinks=true, linkcolor=black, citecolor=blue!50!black, urlcolor=blue!50!black}

\title{StandardE2E: A Unified Framework for End-to-End\\Autonomous Driving Datasets}
\author{
  Stepan Konev\\
  \texttt{stevenkonev@gmail.com}
}
\date{}

\begin{document}
\maketitle

\begin{abstract}
Autonomous driving has shifted from modular perception--prediction--planning stacks toward end-to-end (E2E) models that map sensor inputs directly to vehicle control, often regularized by auxiliary tasks such as 3D detection, motion forecasting, and HD-map perception. Progress is driven by a fast-growing ecosystem of sensor-rich driving datasets, yet each ships its own file formats, APIs, coordinate conventions, and modality coverage, leaving cross-dataset experimentation and even basic per-dataset preprocessing to be re-implemented per project. We present \textsc{StandardE2E}, a framework that provides a single unified interface over E2E driving datasets. \textsc{StandardE2E} (i)~standardizes per-dataset preprocessing under one shared data schema; (ii)~combines multiple datasets in a single PyTorch \texttt{DataLoader} for cross-dataset pretraining, auxiliary-task supervision, and scenario-level filtering; and (iii)~reduces adding a new dataset to a single per-dataset mapping from raw frames to the canonical schema, leaving the entire downstream pipeline unchanged. The framework supports six datasets out of the box --- Waymo End-to-End, Waymo Perception, Argoverse~2 Sensor, Argoverse~2 LiDAR, NAVSIM (OpenScene-v1.1), and WayveScenes101 --- and is released as the open-source \texttt{standard-e2e} Python package, available at \url{https://github.com/stepankonev/StandardE2E}. This document is an extended abstract; a full version with cross-dataset analyses is in preparation.
\end{abstract}

\section{Introduction}
\label{sec:intro}

End-to-end (E2E) autonomous driving has emerged as an alternative to the modular perception--prediction--planning pipeline: a single trainable model maps raw sensor inputs (cameras, LiDAR) directly to vehicle control or planning targets, often regularized by auxiliary supervision on intermediate tasks such as 3D detection, motion forecasting, or HD-map perception~\citep{uniad,vad,paradrive,transfuser,hydramdp}. Progress in this paradigm has been propelled by an expanding ecosystem of large-scale, sensor-rich driving datasets, each released with its own native API, on-disk format, calibration convention, and modality coverage~\citep{wode2e,waymo,argoverse2,navsim,nuplan}.

In practice, this fragmentation imposes a recurring tax on end-to-end driving research. Per project, researchers re-implement preprocessing pipelines from raw protobuf, parquet, pickled-frame, or city-wide GPKG sources into a representation their model can consume. Cross-dataset experimentation --- say, pretraining on Argoverse~2 Sensor before fine-tuning on Waymo End-to-End --- is rare not because the science is uninteresting but because writing $N$ adapters from scratch is expensive and brittle. Existing unification efforts target adjacent problems: \textsc{trajdata}~\citep{ivanovic2023trajdata} provides a unified API for trajectories and vector maps but explicitly excludes sensor data.

\begin{figure}[t]
  \centering
  \includegraphics[width=\linewidth]{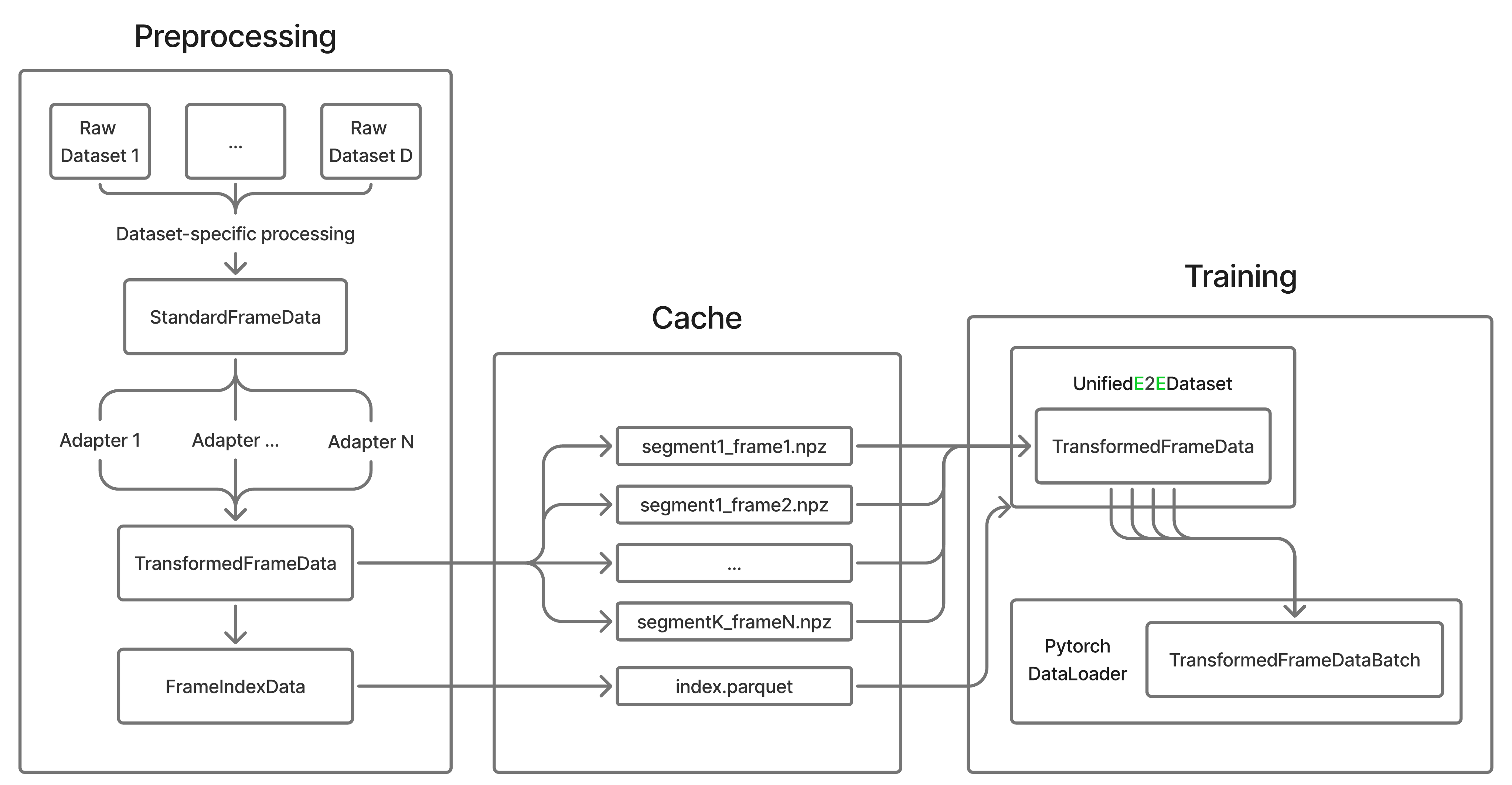}
  \caption{\textsc{StandardE2E} dataflow. The only per-dataset code a contributor writes is the dataset-specific \emph{Raw $\rightarrow$ StandardFrameData} processing (left). All downstream nodes --- the adapter chain, the npz cache, the parquet index, the \texttt{UnifiedE2EDataset}, and the PyTorch \texttt{DataLoader} --- are dataset-agnostic and reused as-is when adding a new dataset.}
  \label{fig:dataflow}
\end{figure}

We address this gap with \textsc{StandardE2E}, a unified preprocessing and loading framework that takes raw E2E driving datasets and produces a shared on-disk representation (Figure~\ref{fig:dataflow}). The central design choice is that \emph{the only per-dataset code a contributor writes is the mapping from raw frames to our canonical \texttt{StandardFrameData} object}; everything downstream --- the adapter chain, indexing, segment-context aggregation, modality defaults, the PyTorch \texttt{Dataset}, and the multi-dataset \texttt{DataLoader} --- is dataset-agnostic and shared across all supported datasets. This minimizes the marginal cost of supporting a new dataset and makes cross-dataset training a single-config-file operation.

\paragraph{Contributions.} (1)~A \textbf{unified preprocessing pipeline} (Figure~\ref{fig:dataflow}) in which a single per-dataset node (\texttt{Raw $\rightarrow$ StandardFrameData}) is the entire user-side surface; the adapter chain, the cache writer and parquet index, segment-context aggregators, modality defaults, scenario filtering, and the \texttt{UnifiedE2EDataset} are reused across datasets. The schema spans cameras, LiDAR (point cloud and BEV), HD maps (vector and BEV), 3D detections, driving command, ego state, and past/future trajectories. (2)~A PyTorch \textbf{\texttt{UnifiedE2EDataset}} that combines multiple datasets in one \texttt{DataLoader}, with per-dataset oversampling, modality whitelisting, scenario-level filtering on the index, and frame-augmentation hooks. (3)~\textbf{Six supported datasets} out of the box --- Waymo End-to-End, Waymo Perception, Argoverse~2 Sensor, Argoverse~2 LiDAR, NAVSIM, and WayveScenes101 --- the last added since our initial release, validating the claim that the per-dataset surface is small.

\section{Related Work}
\label{sec:related}

\paragraph{End-to-end autonomous driving.} A growing line of work trains a single model from sensor inputs to planning or control targets, typically with auxiliary perception losses for regularization: UniAD~\citep{uniad}, VAD~\citep{vad}, ParaDrive~\citep{paradrive}, GenAD~\citep{genad}, TransFuser~\citep{transfuser}, and Hydra-MDP~\citep{hydramdp,hydramdppp}. These models routinely consume four to seven modalities (multi-camera images, LiDAR point clouds or BEVs, HD-map raster or vector input, and ego state), creating downstream demand for unified multimodal data access.

\paragraph{Per-dataset native APIs.} Each major sensor-rich driving dataset ships its own access library: \texttt{waymo-open-dataset} (protobuf-encoded TFRecords)~\citep{waymo}, \texttt{av2-api} (parquet + feather)~\citep{argoverse2}, \texttt{navsim-devkit} on top of \texttt{nuplan-devkit} (pickled per-log scenes plus city-wide GPKG maps)~\citep{navsim,nuplan}, and \texttt{nuscenes-devkit}~\citep{nuscenes}. These define mutually incompatible coordinate conventions, calibration formats, and per-frame schemas, and many keep dataset-wide state in memory at construction time, which is problematic under \texttt{torch.utils.data.DataLoader} multiprocessing.

\paragraph{Unified-format prior work.} The closest precedent is \textsc{trajdata}~\citep{ivanovic2023trajdata}, which provides a unified API to eight pedestrian and driving trajectory datasets with a tabular trajectory representation and a four-element vector-map taxonomy. \textsc{trajdata} scopes itself to trajectories and maps, and its authors flag sensor data as out-of-scope future work. \textsc{StandardE2E} takes up that thread for sensor-rich E2E datasets: we share the two-stage caching philosophy and the goal of a single PyTorch-native \texttt{Dataset}, and we extend the unified schema upward to cover cameras, LiDAR, and a thirteen-channel HD-map taxonomy.

\paragraph{Datasets supported.} At the time of writing, \textsc{StandardE2E} supports six datasets: Waymo End-to-End~\citep{wode2e}, Waymo Perception~\citep{waymo}, Argoverse~2 Sensor~\citep{argoverse2}, Argoverse~2 LiDAR~\citep{argoverse2}, NAVSIM (OpenScene-v1.1)~\citep{navsim}, and WayveScenes101~\citep{wayvescenes}.

\section{The StandardE2E framework}
\label{sec:framework}

\textsc{StandardE2E} centers on a canonical intermediate schema and a single configurable transformation chain that turns raw frames into a training-ready, on-disk cache. A \texttt{StandardFrameData} object represents one frame from one dataset in the canonical format. It carries five mandatory identifiers (\texttt{dataset\_name}, \texttt{split}, \texttt{segment\_id}, \texttt{frame\_id}, \texttt{timestamp}) and a set of optional, Pydantic-validated modality slots: cameras (keyed by \texttt{CameraDirection}), LiDAR, HD map (a list of typed \texttt{MapElement} polylines/polygons), 3D detections, driving command (\texttt{intent}), global position, and past/future ego states, plus free-form \texttt{aux\_data} for dataset-specific metadata. Canonical enums name the modalities (\texttt{Modality}), a thirteen-channel HD-map taxonomy (\texttt{MapElementType}, a strict superset of \textsc{trajdata}'s four), five detection classes (\texttt{DetectionType}), and the columns of the unified \texttt{Trajectory} container. Sections~\ref{sec:chain}--\ref{sec:adding} describe how raw frames flow through this schema to the cache and into a PyTorch \texttt{DataLoader}.

\subsection{The adapter chain}
\label{sec:chain}

The adapter chain is the configurable transformation layer between \texttt{StandardFrameData} and the final \texttt{TransformedFrameData} that is written to disk. It is an \emph{ordered list of adapters}, specified per preprocessing run in a YAML config; each adapter reads the canonical frame, produces or rewrites one modality payload, and passes the frame to the next. Two kinds of adapter make up the chain. \emph{Identity (pass-through) adapters} cache a modality at the source dataset's native fidelity. \emph{Resolution and format adapters} reshape a modality to exactly what the model consumes --- panoramic multi-camera stitching with a configurable maximum image size, point-cloud subsampling, LiDAR bird's-eye-view (BEV) rasterization with a configurable resolution and spatial extent, and HD-map BEV rasterization of the thirteen-channel vector taxonomy with a configurable extent and channel subset.

Conceptually, the chain is the framework's \emph{primary configurability surface}: because the chain alone decides what is computed and written, a user opts into exactly the modalities and resolutions their model requires, and only those land in the cache. Adding a custom transformation does not require modifying the framework: a user-defined adapter subclasses a small \texttt{AbstractAdapter} interface (a single \texttt{transform} method over \texttt{StandardFrameData}) and registers under a string name in the YAML schema, after which the same chain mechanism that ships with the library picks it up. The chain runs once over every raw frame; its output, a \texttt{TransformedFrameData} object, is serialized as one \texttt{npz} file per frame and is exactly what the PyTorch \texttt{Dataset} returns at training time.

\subsection{Segment-context aggregators}
\label{sec:aggregators}

Some E2E supervision signals --- notably past and future ego trajectories and aggregated per-agent detection trajectories --- are often not present in raw per-frame data and must be derived from the surrounding frames in a segment. \textsc{StandardE2E} computes these in a second, post-frame stage of \emph{segment-context aggregators} that runs once after the adapter chain has written every frame in a split. Two aggregators ship with the framework: one composes per-frame ego SE(2) poses into past- and future-trajectory tensors anchored at each frame's local origin, and one aggregates per-agent detection trajectories across the segment. Both are dataset-agnostic, operate on the canonical schema, and fan out across segments via process pools to amortize cost.

\subsection{Indexing and scenario filtering}
\label{sec:indexing}

After preprocessing, every cached frame is described by one row in a parquet index table (\texttt{segment\_id}, \texttt{frame\_id}, \texttt{timestamp}, \texttt{filename}, \texttt{split}, \texttt{dataset\_name}, plus arbitrary extra columns for dataset-specific scenario tags). The index is what the \texttt{UnifiedE2EDataset} sees, and it is what enables scenario-level filtering at training time. A family of composable index filters lets a user define training and validation splits as predicates over the index --- for instance, ``Waymo Perception frames where the ego is moving and a three-second future window exists, plus all NAVSIM frames tagged as urban'' --- without re-running preprocessing, since the cache is shared across runs.

\subsection{The UnifiedE2EDataset}
\label{sec:dataset}

The \texttt{UnifiedE2EDataset} is a PyTorch \texttt{Dataset} that takes the parquet index (possibly the concatenation of per-dataset indices), one or more configured frame loaders that fetch and shape per-frame payloads from the cache, a list of index filters, a list of frame augmentations, and a \texttt{ModalityDefaults} dictionary specifying how to fill modalities a given dataset does not ship. It is fully PyTorch-native and supports the standard \texttt{DataLoader} multiprocessing model; a companion batch class defines collation for variable-length modalities (point clouds, detection sets, trajectories). The same model code can therefore run on frames from several datasets in a single epoch, with modalities a dataset lacks zero-filled by the corresponding default and exposed through a per-modality presence flag.

\subsection{Two-stage architecture and reproducibility}
\label{sec:two-stage}

\textsc{StandardE2E} is structured as two stages. \emph{Stage~1} runs once per dataset and preprocessing config: it materializes the cache by running the adapter chain on every raw frame and writing one npz per frame plus the parquet index. \emph{Stage~2} runs at every training or inference epoch: a flat array load per frame, with no native-API ceremony, no global state, and no per-frame raw-format parsing. The cache is the unit of reproducibility --- the same configuration YAML deterministically produces the same npz layout. Identity adapters preserve modality values bit-for-bit, while resolution and format adapters document their transform explicitly. Per-dataset processor unit tests verify fixture-level equivalence between raw inputs and cached outputs, and a pre-commit pipeline enforces formatting, linting, and type checks on every change.

\subsection{Flexibility as the value proposition}
\label{sec:flexibility}

Because the adapter chain alone determines what is computed and stored, flexibility follows directly, with consequences for both runtime cost and reproducibility. First, \textbf{modality opt-out}: dropping the LiDAR adapter means the cache simply does not carry LiDAR --- no preprocessing time is spent, no disk is consumed, no I/O is performed at training, and no memory is allocated at inference; the same holds for HD-map BEV, segment-context aggregation, and every other modality. Second, \textbf{per-adapter resolution knobs}: camera adapters take a maximum image size; the BEV adapters take a resolution and spatial extent; the HD-map BEV adapter additionally accepts a channel subset. A user who trains on $384$-pixel panoramas and a $32$\,m / $4$-pixel-per-meter BEV pays for exactly that, not the full-resolution multi-sensor payload that native APIs return by default. Native APIs typically support neither.

\subsection{Adding a new dataset}
\label{sec:adding}

The framework's central design claim is that adding a dataset is a localized, well-scoped contribution: the only user-side code is one processor class implementing the \texttt{Raw $\rightarrow$ StandardFrameData} mapping, plus a thin converter that exposes an iterator over raw frames in segment order. The adapter chain, the cache and index writers, the segment-context aggregators, the modality-defaults machinery, and the \texttt{UnifiedE2EDataset} are all reused without modification.

\bibliographystyle{plainnat}
\bibliography{references}

\end{document}